\documentclass[11pt]{article}

\usepackage[final]{acl}

\usepackage{times}
\usepackage{latexsym}

\usepackage[T1]{fontenc}

\usepackage[utf8]{inputenc}

\usepackage{microtype}

\usepackage{inconsolata}

\usepackage{graphicx}

\usepackage{booktabs}
\usepackage{amsmath}
\usepackage{amssymb}
\usepackage{booktabs}
\usepackage{graphicx}
\usepackage{enumitem}
\usepackage{algorithm}
\usepackage{algpseudocode}

\newcommand{\method}{dLLM-KVQ}
\newcommand{\trace}{TRACE}

\title{KV Cache Quantization for Diffusion Language Models}

\author{
 \textbf{Yan Tai\textsuperscript{1,4}},
 \textbf{Yunan Ding\textsuperscript{3,4}},
 \textbf{Guangzhao Li\textsuperscript{1,2}},
 \textbf{Siqi Luo\textsuperscript{1}},
 \textbf{Chaoyang Zhang\textsuperscript{2}},
\\
 \textbf{Guangtao Zhai\textsuperscript{1}},
 \textbf{Guodong Guo\textsuperscript{4,$\dagger$}},
 \textbf{Xiaohong Liu\textsuperscript{1,2,$\dagger$}}
\\
\\
 \textsuperscript{1}Shanghai Jiao Tong University
 \textsuperscript{2}Shanghai Innovation Institute
\\
 \textsuperscript{3}The Hong Kong Polytechnic University
\\
 \textsuperscript{4}Ningbo Institute of Digital Twin, Eastern Institute of Technology, Ningbo, China
\\
\\
 \small{
   \textbf{Correspondence:} \href{mailto:gdguo@eitech.edu.cn}{gdguo@eitech.edu.cn}, \href{mailto:xiaohongliu@sjtu.edu.cn}{xiaohongliu@sjtu.edu.cn}
 }
}

\begin{document}
\maketitle

\begin{abstract}
KV cache quantization is well studied for autoregressive language models, where cached prefix states are exact under
causal attention and quantization is mainly a cache reconstruction problem. Diffusion language models (dLLMs) break
this assumption. Fast dLLM inference reuses KV states across denoising steps, but bidirectional token representations
can change as masked tokens are revealed, so the reused cache may already differ from full recomputation before
quantization is applied. We study KV cache quantization under this cache-reuse error. We adapt KIVI and KVQuant to
Fast-dLLM-style prefix and dual cache reuse, and introduce \trace{} and \method{}. \trace{} reconstructs feasible dLLM
reveal trajectories from prompt-answer pairs, providing calibration states that match iterative masked decoding.
\method{} uses these trajectories to choose per-layer K-cache entries that remain in high precision according to
output- and attention-level behavior metrics, while applying standard low-bit value quantization. Across LLaDA and
Dream evaluations on GSM8K, HumanEval, MATH, and MBPP, \method{} achieves the highest average score among quantized methods across
the reported 4-bit and 2-bit prefix/dual cache settings. 
\end{abstract}

\section{Introduction}
\label{sec:introduction}

Diffusion language models (dLLMs) generate text by iteratively refining partially masked sequences rather than by
strict left-to-right next-token prediction~\citep{austin2021structured,lou2023discrete,sahoo2024simple,shi2024simplified,nie2026large,ye2025dream}. Efficient dLLM inference therefore often reuses KV states across denoising steps, as in the prefix and dual cache modes of Fast-dLLM~\citep{wu2025fast} and adaptive cache updates in dLLM-Cache~\citep{liu2025dllm}. This makes KV cache compression a natural next step, but it also changes the assumptions behind standard autoregressive (AR) KV cache quantization.

\begin{figure*}[t]
  \centering
  \includegraphics[width=\textwidth]{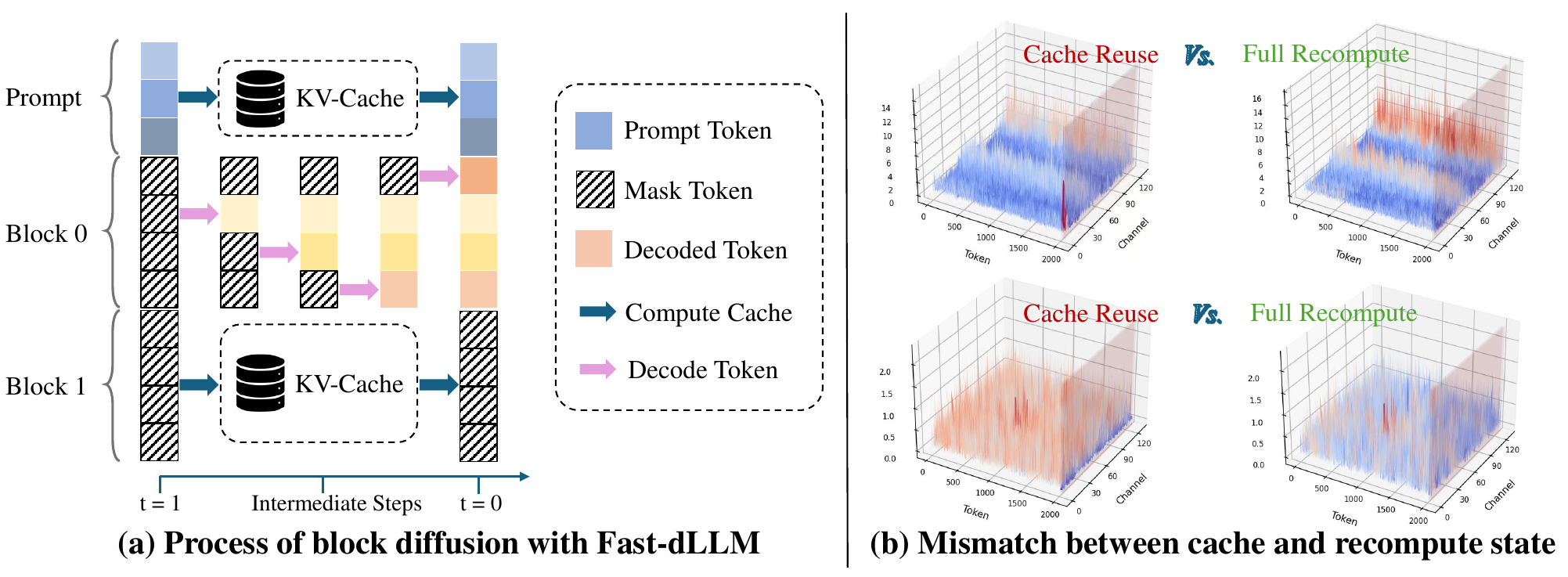}
  \caption{Fast-dLLM cache reuse and cache-reuse error. (a) Dual cache reuses prefix and suffix KV states from a
  block-initial forward pass while recomputing only the active block. (b) The reused cache can differ from the
  corresponding full-recompute cache after masked tokens inside the block are revealed.}
  \label{fig:cache-reuse-error}
\end{figure*}

In AR decoding, cached prefix states are exact under causal self-attention~\citep{vaswani2017attention}: future
generated tokens do not change the key and value states of previous tokens. In dLLM decoding, layers use bidirectional
self-attention over the current masked sequence. When tokens inside the active block are revealed, prefix or suffix
token states can also change under full recomputation. Cache reuse avoids this recomputation by keeping selected
regions fixed, so the reused cache is an approximation to the full-recompute cache. Figure~\ref{fig:cache-reuse-error} (b)
illustrates this difference.

This difference makes direct AR-to-dLLM transfer unreliable. AR KV quantizers such as KIVI~\citep{liu2024kivi} and
KVQuant~\citep{hooper2024kvquant} are usually motivated as approximating an unquantized reference cache with fewer
bits. In a dLLM cache-reuse path, quantization is applied to a cache that may already differ from full recomputation.
The deployed cache therefore contains both cache-reuse error and quantization error. Reducing reconstruction error of
the reused cache alone may not preserve attention patterns, token scores, or final task decisions.

This setting creates two practical challenges. First, calibration data must reflect masked denoising states rather than
only complete prompt-answer sequences, because the same final answer can be reached through different reveal orders.
Second, the quantization objective should be tied to model behavior. This is especially important for K caches, where a
small numerical perturbation can shift attention mass among reused and recomputed tokens.

To our knowledge, this is the first systematic study of KV cache quantization for dLLMs. We adapt KIVI and KVQuant to a
unified Fast-dLLM cache interface and evaluate them under prefix and dual cache reuse. We then introduce \trace{}, a
trajectory reconstruction procedure that assigns each answer token in a prompt-answer pair to a legal reveal step under
the dLLM block schedule. These reconstructed denoising states provide calibration data that matches masked iterative
decoding without storing full multi-step logits or KV caches. Figure~\ref{fig:trace-calibration} summarizes how this
trajectory construction supports offline policy fitting and online cache quantization.

Building on this calibration, \method{} selects a small set of K-cache entries to keep in high precision. The policy is
chosen per model, cache mode, bit-width, and layer using output- and attention-level metrics measured on reconstructed
trajectories. At inference time, it uses only information available from the block-initial forward pass, while the V
cache follows a standard low-bit quantize-dequantize path. Across LLaDA-Instruct and Dream-Base-7B on GSM8K, HumanEval, MATH, and
MBPP, \method{} obtains the highest average score among the evaluated quantized methods in the reported 4-bit and
2-bit prefix/dual settings.


The paper makes four contributions:
\begin{itemize}[leftmargin=*, itemsep=2pt, topsep=2pt, parsep=0pt, partopsep=0pt]
  \item We formulate dLLM KV cache quantization under cache reuse and separate cache-reuse error from quantization error.
  \item We propose \trace{} to reconstruct dLLM reveal trajectories from prompt-answer pairs for calibration.
  \item We propose \method{} to select per-layer high-precision K-token budgets from behavior metrics and apply them using only block-initial information.
  \item We show that \method{} obtains the best four-task average among quantized methods across all reported models, cache modes, and bit-widths.
\end{itemize}
\section{Preliminary}
\label{sec:preliminary}

\paragraph{dLLM Block Diffusion.} Following discrete and masked diffusion language modeling
\citep{austin2021structured,lou2023discrete,sahoo2024simple,shi2024simplified,arriola2025block}, let $p$ be the prompt and $x_t$ be the partially generated response at denoising step $t$, where unrevealed positions
are \texttt{<mask>}. Unlike AR decoding, which appends one new token to a fixed prefix, a dLLM repeatedly rewrites a
partially masked sequence. In block-wise decoding, the response is divided into blocks $\mathcal{B}_1,\ldots,
\mathcal{B}_B$. At each denoising step $t$ inside the current block $\mathcal{B}_b$, the model predicts
distributions for masked positions and
reveals a scheduled subset of high-confidence tokens, following the iterative masked decoding pattern used in
non-autoregressive and diffusion-style generation~\citep{ghazvininejad2019mask,wu2025fast}:
\[
  x_{t+1,i} =
  \begin{cases}
    \arg\max_v p_\theta(v \mid p,x_t,i), & i \in \mathcal{R}_{b,t},\\
    x_{t,i}, & \text{otherwise},
  \end{cases}
\]
where $\mathcal{R}_{b,t}\subseteq \mathcal{B}_b$ is the set of positions revealed at step $t$. 

\paragraph{dLLM Cache.} Following Fast-dLLM~\citep{wu2025fast} and recent adaptive dLLM caching
work~\citep{liu2025dllm}, we consider two cache reuse modes for efficient dLLM inference. If the model is fully recomputed at every denoising step, layer $\ell$ produces a full cache
$C_t^\ell=(K_t^\ell,V_t^\ell)$ from the current state $(p,x_t)$. Fast-dLLM avoids this cost by running a full forward
pass at the beginning of a block, storing the resulting KV cache, and reusing selected regions during later denoising
steps. Prefix cache reuses the prefix
region and recomputes the active block plus suffix. Dual cache reuses both prefix and suffix and recomputes only the
active block. We denote either cache mode by
\[
  \begin{aligned}
  C_{t,m}^{\ell,\mathrm{reuse}}
  &=
  \Gamma_m\!\left(C_{\tau_b}^{\ell,\mathrm{full}},
                  C_t^{\ell,\mathrm{local}}, \mathcal{B}_b\right),\\
  m&\in\{\mathrm{prefix},\mathrm{dual}\}.
  \end{aligned}
\]
where $\tau_b$ is the block-initial step at which the reusable cache is built,
$C_t^{\ell,\mathrm{local}}$ is the cache recomputed for the active regions, and $\Gamma_m$ selects which token regions
are reused and which are recomputed. Figure~\ref{fig:cache-reuse-error}(a) illustrates Fast-dLLM's dual cache reuse process, where cached regions (prefix and suffix) are reused while the active block is progressively denoised over intermediate steps.
\section{Why dLLM KV Cache Quantization Is Different}
\label{sec:analysis}

KV cache quantization for autoregressive (AR) LLMs is usually framed as a reconstruction problem
\citep{liu2024kivi,hooper2024kvquant}: the unquantized cache is the reference cache, and the quantizer should
approximate it with less memory. For diffusion language models
(dLLMs), this view is incomplete. dLLM cache methods such as Fast-dLLM accelerate dLLM decoding by reusing cache states
computed by a block-initial forward pass, which are approximations to full recomputation~\citep{wu2025fast,liu2025dllm}. Our quantization study is applied on top of this cache
reuse mechanism. Therefore, the cache to be quantized may already differ from the cache that full recomputation would
produce. This section defines this difference and explains why it changes the quantization problem.

\subsection{Cache-Reuse Error}
\label{subsec:cache-reuse-error}

The key difference from AR decoding is that dLLM layers use bidirectional self-attention over the current masked
sequence. Consider a block after its initial step $\tau_b$, and a later denoising step $t$ in the same block. Some
positions in $\mathcal{B}_b$ may have changed from \texttt{<mask>} to concrete tokens during these intervening
steps. Under full recomputation, prefix and suffix token states are recomputed together with this updated block, so
their K/V entries can change. Under prefix or dual cache reuse, the reused prefix or suffix entries remain the ones
computed at $\tau_b$.

We write this cache-reuse error as:
\[
  E_{t,m}^{\ell,\mathrm{reuse}}
  =
  C_{t,m}^{\ell,\mathrm{reuse}} - C_t^{\ell,\mathrm{full}}.
\]
In AR LLMs, the analogous term is zero for cached prefix tokens under standard causal attention~\citep{vaswani2017attention}: past token states do
not depend on future generated tokens. In dLLMs, $E^{\mathrm{reuse}}$ is generally nonzero. The key and value plots in
Figure~\ref{fig:cache-reuse-error} make this empirical fact visible: even for the same block, layer, and head, the reused
cache state and the later full-recompute cache can have different feature distributions.

\subsection{Quantization Error Is Not the Only Error}
\label{subsec:dllm-quant-error}

For AR cache quantization, the deployed cache is usually $\widehat{C}_t=Q_\beta(C_t)$, where $Q_\beta$ is a
$\beta$-bit quantizer
\citep{liu2024kivi,hooper2024kvquant,lin2025qserve}.
The cache error is simply $Q_\beta(C_t)-C_t$, so improving cache reconstruction is a natural objective.

For dLLMs with cache reuse, quantization is applied to $C_{t,m}^{\mathrm{reuse}}$, but the reference behavior we
would like to preserve is the behavior of $C_t^{\mathrm{full}}$. The resulting cache error is therefore
\[
  \begin{aligned}
  \widehat{C}_{t,m}^{\ell} - C_t^{\ell,\mathrm{full}}
  &=
  \underbrace{C_{t,m}^{\ell,\mathrm{reuse}} - C_t^{\ell,\mathrm{full}}}_{\text{cache-reuse error}}
  \\
  &\quad+
  \underbrace{Q_\beta(C_{t,m}^{\ell,\mathrm{reuse}})
  -C_{t,m}^{\ell,\mathrm{reuse}}}_{\text{quantization error}}.
  \end{aligned}
\]
This decomposition is the main reason AR reconstruction objectives are insufficient here. A dLLM quantizer is not only
approximating an unquantized cache; it is modifying a cache that may already differ from full recomputation. As a
result, reducing quantization error alone may fail to predict generation quality.


\subsection{Problem Definition}
\label{subsec:problem-definition}

Given a distribution of denoising states $\mathcal{T}$, a cache mode $m$, and a memory budget $\rho$, the goal is to
learn or choose an online cache quantization policy that preserves model behavior under cache reuse. Let
$y_t^{\mathrm{full}}=f_\theta(p,x_t,C_t^{\mathrm{full}})$ and
$\widehat{y}_{t,m}=f_\theta(p,x_t,Q_\beta(C_{t,m}^{\mathrm{reuse}}))$. We write the objective as
\[
  \begin{aligned}
  \min_{Q_\beta}\quad
  &
  \mathbb{E}_{(p,x_t,m)\sim\mathcal{T}}
  \left[
    \mathcal{D}_{\mathrm{beh}}(y_t^{\mathrm{full}}, \widehat{y}_{t,m})
  \right]
  \\
  \text{s.t.}\quad
  &
  \operatorname{Mem}(Q_\beta(C)) \leq \rho\,\operatorname{Mem}(C).
  \end{aligned}
\]
Here $\mathcal{D}_{\mathrm{beh}}$ can measure output logits, token margins, attention behavior, or downstream
decisions. The next challenge is calibration: representative denoising states are needed, but storing full multi-step
logits and KV caches is too expensive. The following method section addresses this with \trace{} trajectory
calibration and the \method{} quantization policy.

\section{\method}
\label{sec:method}

Section~\ref{sec:analysis} shows that dLLM cache quantization must handle cache-reuse error in addition to ordinary
quantization error. \method{} addresses this with a calibration-and-deployment design. The offline stage reconstructs
likely dLLM decoding trajectories and uses them to choose which K-cache entries stay in high precision. The online stage then
applies this policy to the cache states produced by the block-initial forward pass and reused by the cache mechanism.
Thus full recomputation is used only for calibration analysis, not during deployment.
Figure~\ref{fig:trace-calibration} shows the full pipeline.

\paragraph{K-side focus.}
The design goal is behavior preservation under cache reuse, especially stable attention choices. K states enter the
query-key scores that define attention weights, while V states are read only after those weights are chosen. K
quantization can therefore redirect attention across reused and recomputed positions; V quantization perturbs the
content at positions already selected by attention. \method{} accordingly spends the adaptive high-precision budget on
K and keeps V on a conservative low-bit QDQ path.

\subsection{\trace{} Trajectory Calibration}
\label{subsec:trace}

\begin{figure*}[t]
  \centering
  \includegraphics[width=\textwidth]{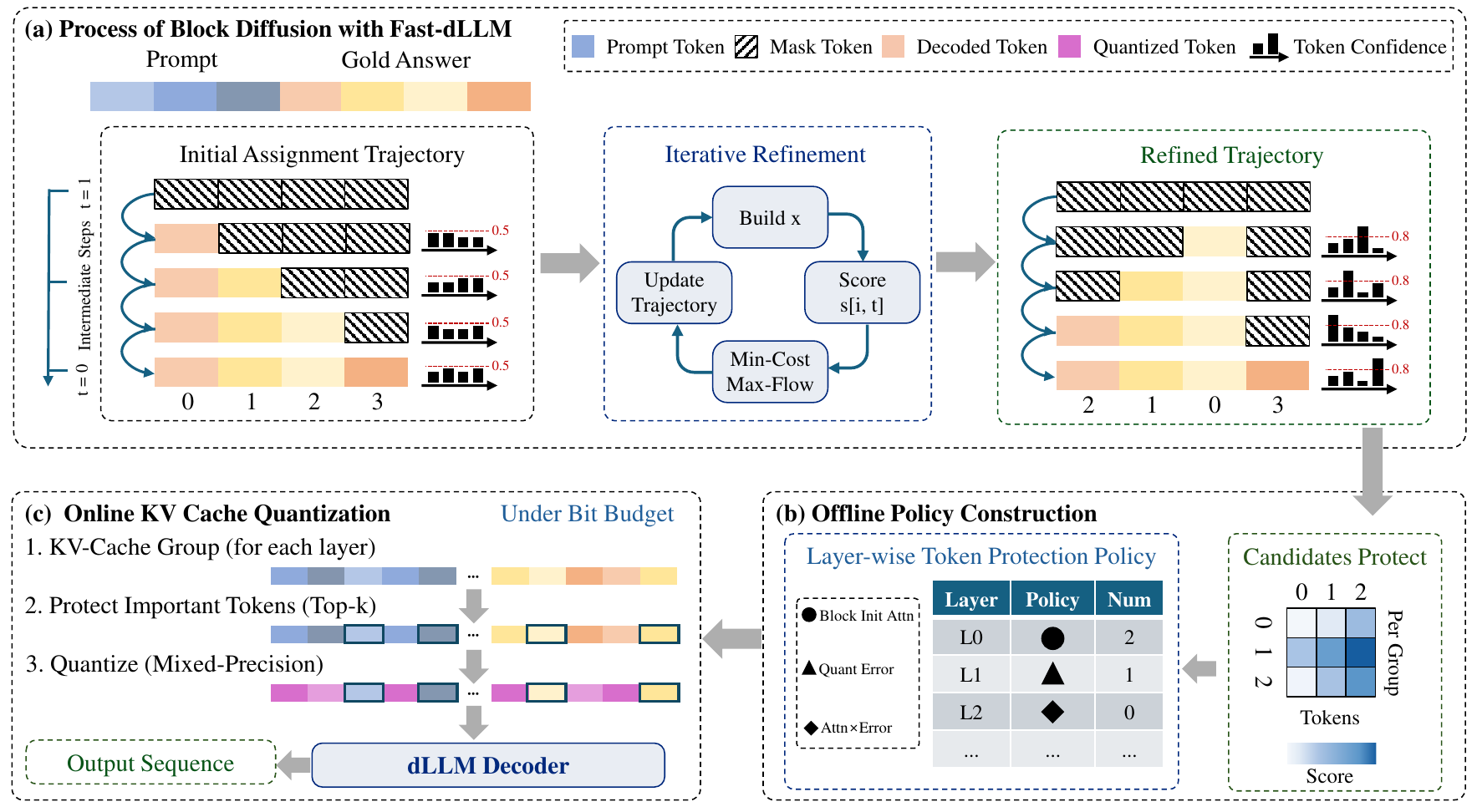}
  \caption{\method{} pipeline. (a) \trace{} reconstructs masked-denoising trajectories from prompt-answer pairs.
  (b) Offline diagnostics fit per-layer K-cache policies from behavior and attention metrics. (c) Online quantization
  applies the selected policy to block-initial reused cache states.}
  \label{fig:trace-calibration}
\end{figure*}

\trace{} is designed to answer a simple calibration question: given a prompt and a final answer, which reveal trajectory
is most consistent with the model under the dLLM block schedule? This differs from standard AR calibration, where a
complete sequence can be processed once. A dLLM calibration sample must instead contain intermediate masked states,
because the model's prediction and cache behavior depend on which answer tokens have already been revealed.
Figure~\ref{fig:trace-calibration}(a) illustrates this trajectory construction stage.

For a prompt-answer pair $(p,y)$, \trace{} assigns each answer token $y_i$ to one legal reveal step. The assignment is
initialized by any schedule-consistent allocation, i.e., every token is placed inside its block window and each step
receives the prescribed reveal budget. For a token $y_i$ in block $b$, the legal window $\mathcal{W}_i$ is the set of
reveal steps assigned to that block. The reveal budgets $\{k_t\}$ satisfy $\sum_t k_t=|y|$ and define how many answer
tokens are assigned to each step in calibration. Given an assignment at refinement round $r$, we roll out the
corresponding masked states $\{x_t^{(r)}\}$ and compute the state matrix
$s^{(r)}_{i,t}=\log p_\theta(y_i\mid p,x_t^{(r)})$ for legal token-step pairs. The next assignment is obtained by
\[
  A^{(r+1)}
  =
  \arg\max_{A\in\mathcal{C}}
  \langle A,S^{(r)}\rangle ,
  \label{eq:trace-refinement}
\]
where $\mathcal{C}$ is the set of assignments satisfying the block window and per-step reveal-budget constraints. This
linear assignment problem is solved as min-cost max-flow~\citep{edmonds1972theoretical} with edge cost $-s^{(r)}_{i,t}$. Equivalently, each answer
token sends one unit of flow to one legal step, and each step receives exactly $k_t$ units. The rollout and assignment
steps are repeated until the trajectory stabilizes or a small refinement limit is reached. The saved artifact contains
only the prompt, answer, and reveal assignment. Logits and KV caches are recomputed on demand for later diagnostics.


\begin{algorithm}[t]
\small
\caption{\trace{} trajectory reconstruction}
\label{alg:trace-pseudocode}
\begin{algorithmic}[1]
\Require prompt $p$, answer $y$, reveal budgets $\{k_t\}$, legal windows $\{\mathcal{W}_i\}$, rounds $R$
\State Initialize a feasible assignment $A^{(0)}\in\mathcal{C}$
\For{$r=0,\ldots,R-1$}
    \State Build masked states $\{x_t^{(r)}\}$ from $A^{(r)}$
    \State Compute $S^{(r)}=[\log p_\theta(y_i\mid p,x_t^{(r)})]$
    \State Solve Eq.~\ref{eq:trace-refinement} by min-cost flow
    \If{$A'=A^{(r)}$}
        \State \Return $A'$
    \EndIf
    \State Set $A^{(r+1)}\leftarrow A'$
\EndFor
\State \Return $A^{(R)}$
\end{algorithmic}
\end{algorithm}

\paragraph{Quantization format.}
We use quantize-dequantize (QDQ) to denote the simulated low-bit cache path, a common abstraction in post-training
quantization and low-bit LLM serving~\citep{yao2022zeroquant,xiao2023smoothquant,shao2024omniquant,lin2025qserve}.
\method{} keeps the K/V grouping used by representative AR KV-cache quantizers~\citep{liu2024kivi,hooper2024kvquant}; the
implementation details and distribution plots are given in Appendix~\ref{app:quant-granularity}.

\subsection{K-Cache Policy Selection}
\label{subsec:offline-policy}

The policy determines how many K-cache tokens in each group should remain in high precision instead of being quantized
to low bit-width. The use of
attention-derived token scores is related in spirit to AR KV retention methods that exploit persistent or prompt-local
attention concentration~\citep{zhang2023h2o,xiao2024efficient,li2024snapkv,cai2024pyramidkv}, but here the score is
used to decide which K tokens keep high precision rather than which tokens remain in the cache. For a cache
mode $m$ and layer $\ell$, the policy selects a candidate $a=(s,p)$: $s$ is a token-importance score and $p$ is the
number of K tokens per group kept in high precision. The candidate score can be based on attention from the
block-initial forward pass, K quantization error when all K tokens are quantized, or their normalized product. The
high-precision budget is selected from $\{0,1,2,4\}$ in our main experiments. Budget $p=0$ corresponds to ordinary
low-bit K quantization.

The full-recompute cache in Section~\ref{sec:analysis} is the reference used to define cache-reuse error. Policy
selection is instead expressed relative to the same reused cache with all K tokens quantized, because every
inference-time candidate starts from that cache state. This offline fitting stage is summarized in
Figure~\ref{fig:trace-calibration}(b). For a record $r$ from a reconstructed trajectory--identified by
a calibration sample, cache mode, block, denoising step, and layer--and candidate $a$, let $\Delta_q(a,r)$ be the
improvement over this baseline for behavior metric $q$. Metrics whose smaller values are better, such as score
distance or attention KL, use baseline minus candidate. Metrics whose larger values are better, such as top-1
agreement or top-$K$ recall, use candidate minus baseline. In our experiments, $q$ includes centered score distance,
attention KL, top-1 key agreement, and top-$K$ key recall. Centered score distance compares the logit-score vector
after removing its mean; the key-agreement metrics test whether the largest-attended or top-$K$ attended cache tokens
are preserved. Define
\[
  \begin{aligned}
  \mathcal{L}_{m,\ell}(a)
  &=
    -\sum_q w_q\,
    \mathbb{E}_{r\in\mathcal{D}_{m,\ell}}\Delta_q(a,r)\\
  &\quad+
    \lambda\,\mathbb{E}_{r\in\mathcal{D}_{m,\ell}}B(a,r),
  \end{aligned}
  \label{eq:policy-objective}
\]
Here $\mathcal{D}_{m,\ell}$ is the diagnostic record set for cache mode $m$ and layer $\ell$, $w_q$ are fixed metric
weights, $\lambda$ weights the bit-width penalty, and $B(a,r)$ measures effective bit-width. \method{} selects
$a^*_{m,\ell}=\arg\min_{a\in\mathcal{A}_{m,\ell}}\mathcal{L}_{m,\ell}(a)$. The feasible set also enforces reliability
constraints, including a maximum effective K bit-width and a minimum win rate on the score-distance metric. Policies
are selected separately for each model, cache mode, target bit-width, and layer. This objective is not optimized by
gradient descent. It chooses a discrete deployment policy from trajectory-level measurements.

\subsection{Online \method{} Quantization}
\label{subsec:online-inference}

At inference time, \method{} uses only information available during the block-initial forward pass. For the
current cache mode and layer, it retrieves the selected score type and high-precision budget, as shown in
Figure~\ref{fig:trace-calibration}(c). If the score is based on
attention, the importance of cache token $i$ is the average attention probability, computed in that block-initial pass,
assigned by active-block queries:
\[
  \begin{aligned}
  I_i^{\mathrm{attn}}
  &=
    \frac{1}{|\mathcal{H}|\,|\mathcal{Q}_b|}
    \sum_{h\in\mathcal{H}}\sum_{q\in\mathcal{Q}_b}
    A^{\mathrm{init}}_{\ell,h,q,i},\\
  A^{\mathrm{init}}_{\ell,h,q,i}
  &=
    \left[
    \operatorname{softmax}
    \left(
    \frac{q^{\mathrm{init}}_{\ell,h,q}(K^{\mathrm{init}}_{\ell,h})^\top}{\sqrt{d_h}}
    \right)
    \right]_i .
  \end{aligned}
  \label{eq:reuse-attn-importance}
\]
Here $\mathcal{H}$ is the set of attention heads, $\mathcal{Q}_b$ is the set of query positions in the active block
$\mathcal{B}_b$, and $d_h$ is the head dimension.
If the score is quantization error, the importance is the token-wise reconstruction error produced by ordinary
low-bit K quantization without protection:
\[
  I_i^{\mathrm{qe}}
  =
  \left\|
  K^{\mathrm{init}}_{\ell,i}
  -
  \left[\operatorname{QDQ}^{\mathrm{base}}_\beta(K^{\mathrm{init}}_{\ell,G(i)})\right]_i
  \right\|_2 .
  \label{eq:k-quant-error-importance}
\]
Here $\operatorname{QDQ}^{\mathrm{base}}_\beta$ denotes the baseline K quantize-dequantize operator at bit-width
$\beta$ with
high-precision budget $p=0$, and $G(i)$ is the token group containing token $i$. The hybrid score multiplies normalized
block-initial attention and normalized quantization error.

KIVI-style K quantization groups K-cache tokens along the token dimension. Within each group, \method{} keeps the
top-$p$ tokens under $I_i$ in high precision. The remaining tokens are quantized with min-max quantize-dequantize, where the range is
estimated only from the tokens that will be quantized in the same group. This exclusion prevents high-precision outliers
from expanding the quantization grid for the low-bit tokens. The V cache uses the standard contiguous low-bit quantize-dequantize described in
Appendix~\ref{app:quant-granularity}, without residual tails or clipping in the submitted method.

The offline and online stages are deliberately separated. Offline diagnostics may compare candidate policies against
full recomputation to choose the policy. Online inference does not access full recomputation. It applies the
selected policy to the reused cache state already required by dLLM cache reuse.

\section{Experiments}
\label{sec:experiments}

\paragraph{Setup.}
Our experimental protocol follows Fast-dLLM~\citep{wu2025fast}, which provides the prefix and dual cache-reuse mechanisms studied in this
work. We evaluate on LLaDA-Instruct~\citep{nie2026large} and Dream-Base-7B~\citep{ye2025dream} with the same benchmark
suite: GSM8K~\citep{cobbe2021gsm8k}, HumanEval~\citep{chen2021humaneval}, MATH~\citep{hendrycks2021measuring}, and
MBPP~\citep{austin2021mbpp}.
All reported runs use generation length 256, block length 32, and Fast-dLLM's confidence-threshold parallel decoding
with threshold 0.9. We evaluate 4-bit and 2-bit KV-cache quantization in both prefix and dual cache modes, using NVIDIA
H200 140GB GPUs. Detailed few-shot settings, calibration data construction, and quantizer hyperparameters are provided
in Appendix~\ref{app:experimental-setup}.

\paragraph{Baselines.}
The no-cache row is the original-inference reference. The direct baselines for cache quantization are the prefix-cache
and dual-cache rows, because quantized methods are applied on top of these cache mechanisms. We compare
against KIVI~\citep{liu2024kivi} and KVQuant~\citep{hooper2024kvquant} after adapting both methods to the same dLLM cache interface.

\begin{table*}[t]
\centering
\small
\setlength{\tabcolsep}{3pt}

\begin{minipage}[t]{0.49\textwidth}
\centering
\textbf{(a) 4-bit}
\vspace{2pt}

\resizebox{\linewidth}{!}{
\begin{tabular}{lrrrrr}
\toprule
Method & GSM8K & HumanEval & MATH & MBPP & Avg. \\
\midrule
LLaDA full attention & 78.62 & 42.07 & 33.36 & 29.80 & 45.96 \\
\midrule
\multicolumn{6}{l}{\textit{Prefix cache}} \\
Cache only & 78.47 & 42.07 & 33.18 & 26.60 & 45.08 \\
\cmidrule(lr){1-6}
KIVI & 77.33 & 42.68 & 32.40 & 25.60 & 44.50 \\
KVQuant & 77.48 & 42.07 & 32.68 & 24.60 & 44.21 \\
\method{} & \textbf{77.94} & \textbf{43.90} & \textbf{32.92} & \textbf{26.60} & \textbf{45.34} \\
\midrule
\multicolumn{6}{l}{\textit{Dual cache}} \\
Cache only & 77.94 & 35.98 & 32.62 & 25.80 & 43.09 \\
\cmidrule(lr){1-6}
KIVI & 76.35 & \textbf{37.20} & 32.62 & 25.60 & 42.94 \\
KVQuant & 78.01 & 35.37 & 32.90 & 25.80 & 43.02 \\
\method{} & \textbf{78.62} & 34.76 & \textbf{33.02} & \textbf{26.80} & \textbf{43.30} \\
\bottomrule
\end{tabular}
}
\end{minipage}
\hfill
\begin{minipage}[t]{0.49\textwidth}
\centering
\textbf{(b) 2-bit}
\vspace{2pt}

\resizebox{\linewidth}{!}{
\begin{tabular}{lrrrrr}
\toprule
Method & GSM8K & HumanEval & MATH & MBPP & Avg. \\
\midrule
LLaDA full attention & 78.62 & 42.07 & 33.36 & 29.80 & 45.96 \\
\midrule
\multicolumn{6}{l}{\textit{Prefix cache}} \\
Cache only & 78.47 & 42.07 & 33.18 & 26.60 & 45.08 \\
\cmidrule(lr){1-6}
KIVI & 76.50 & 38.41 & 32.38 & 27.20 & 43.62 \\
KVQuant & 76.27 & \textbf{39.63} & 32.30 & 17.60 & 41.45 \\
\method{} & \textbf{77.18} & \textbf{39.63} & \textbf{32.90} & \textbf{27.80} & \textbf{44.38} \\
\midrule
\multicolumn{6}{l}{\textit{Dual cache}} \\
Cache only & 77.94 & 35.98 & 32.62 & 25.80 & 43.09 \\
\cmidrule(lr){1-6}
KIVI & 77.56 & 34.15 & 31.86 & \textbf{27.80} & 42.84 \\
KVQuant & 77.48 & \textbf{35.98} & 31.94 & 17.00 & 40.60 \\
\method{} & \textbf{77.79} & \textbf{35.98} & \textbf{32.04} & 27.20 & \textbf{43.25} \\
\bottomrule
\end{tabular}
}
\end{minipage}

\caption{LLaDA results under 4-bit and 2-bit KV cache quantization. LLaDA full attention denotes no-cache original inference. Cache-only rows are unquantized cache references. For \method{}, each row reports the selected configuration in the current result matrix. Average denotes the arithmetic mean over GSM8K, HumanEval, MATH, and MBPP.}
\label{tab:llada-gen256-quant}
\end{table*}

\begin{table*}[t]
\centering
\small
\setlength{\tabcolsep}{3pt}

\begin{minipage}[t]{0.49\textwidth}
\centering
\textbf{(a) 4-bit}
\vspace{2pt}

\resizebox{\linewidth}{!}{
\begin{tabular}{lrrrrr}
\toprule
Method & GSM8K & HumanEval & MATH & MBPP & Avg. \\
\midrule
Dream full attention & 74.15 & 39.63 & 34.18 & 54.40 & 50.59 \\
\midrule
\multicolumn{6}{l}{\textit{Prefix cache}} \\
Cache only & 74.60 & 43.29 & 34.20 & 55.20 & 51.82 \\
\cmidrule(lr){1-6}
KIVI & 73.69 & 40.85 & 33.74 & 54.20 & 50.62 \\
KVQuant & \textbf{74.83} & 37.80 & 33.86 & 54.60 & 50.27 \\
\method{} & 74.53 & \textbf{41.46} & \textbf{34.28} & \textbf{56.20} & \textbf{51.62} \\
\midrule
\multicolumn{6}{l}{\textit{Dual cache}} \\
Cache only & 74.91 & 37.20 & 33.50 & 53.00 & 49.65 \\
\cmidrule(lr){1-6}
KIVI & 73.24 & 37.20 & 33.70 & 53.00 & 49.29 \\
KVQuant & \textbf{75.89} & 35.98 & 33.30 & 52.80 & 49.49 \\
\method{} & 74.91 & \textbf{37.80} & \textbf{34.04} & \textbf{53.60} & \textbf{50.09} \\
\bottomrule
\end{tabular}
}
\end{minipage}
\hfill
\begin{minipage}[t]{0.49\textwidth}
\centering
\textbf{(b) 2-bit}
\vspace{2pt}

\resizebox{\linewidth}{!}{
\begin{tabular}{lrrrrr}
\toprule
Method & GSM8K & HumanEval & MATH & MBPP & Avg. \\
\midrule
Dream full attention & 74.15 & 39.63 & 34.18 & 54.40 & 50.59 \\
\midrule
\multicolumn{6}{l}{\textit{Prefix cache}} \\
Cache only & 74.60 & 43.29 & 34.20 & 55.20 & 51.82 \\
\cmidrule(lr){1-6}
KIVI & 71.57 & 40.24 & 31.52 & 52.60 & 48.98 \\
KVQuant & \textbf{73.92} & 36.59 & 32.28 & 52.20 & 48.75 \\
\method{} & 71.80 & \textbf{42.07} & \textbf{32.36} & \textbf{53.20} & \textbf{49.86} \\
\midrule
\multicolumn{6}{l}{\textit{Dual cache}} \\
Cache only & 74.91 & 37.20 & 33.50 & 53.00 & 49.65 \\
\cmidrule(lr){1-6}
KIVI & 70.43 & \textbf{39.02} & 31.36 & 51.60 & 48.10 \\
KVQuant & \textbf{72.71} & 29.88 & 31.52 & 51.20 & 46.33 \\
\method{} & 71.72 & \textbf{39.02} & \textbf{32.56} & \textbf{52.00} & \textbf{48.83} \\
\bottomrule
\end{tabular}
}
\end{minipage}

\caption{Dream results under 4-bit and 2-bit KV cache quantization. Dream full attention denotes no-cache original inference. Cache-only rows are unquantized cache references. For \method{}, each row reports the selected configuration in the current result matrix. Average denotes the arithmetic mean over GSM8K, HumanEval, MATH, and MBPP.}
\label{tab:dream-gen256-quant}
\end{table*}

\subsection{Main Results}
\label{subsec:main-results}

Tables~\ref{tab:llada-gen256-quant} and~\ref{tab:dream-gen256-quant} report the main result matrix. Each quantized row
should be compared with the matching cache-only reference in the same model, cache mode, and bit-width block. Across
both models, both cache modes, and both bit-widths, \method{} obtains the highest average score among the evaluated
quantized methods.

For LLaDA, \method{} is strongest in the prefix-cache setting. At 4-bit it improves the average over KIVI and KVQuant
(45.34 versus 44.50 and 44.21) and also recovers the cache-only MBPP score. At 2-bit, where ordinary low-bit cache
quantization is more fragile, \method{} keeps the best average and avoids the large MBPP drop observed for KVQuant.
Dual cache is more mixed at the task level: KIVI is better on HumanEval in the 4-bit setting and KIVI or KVQuant tie
\method{} on some 2-bit tasks. However, \method{} still gives the best dual-cache average at both bit-widths, mainly by
keeping GSM8K, MATH, and MBPP from moving in different adverse directions.

Dream shows a different failure pattern. KVQuant is often the strongest method on GSM8K, especially in dual cache, but
the same rows lose accuracy on HumanEval, MATH, and MBPP. This task imbalance is why KVQuant has lower average scores
than \method{} in every Dream setting, even when it wins one task. \method{} does not dominate every individual cell;
rather, it gives the most stable average behavior across tasks, models, cache modes, and bit-widths. This supports the
analysis in Section~\ref{sec:analysis}: cache reconstruction error alone is not a reliable predictor of generation
quality under dLLM cache reuse, because the useful quantization policy depends on how cache perturbations affect
attention and output decisions.

\subsection{Ablation Studies}
\label{subsec:ablations}

The ablations use LLaDA, MBPP, 4-bit quantization, and Fast-dLLM cache modes to isolate whether the selected K policy is
useful at inference time.

\begin{table*}[t]
\centering
\small
\setlength{\tabcolsep}{5pt}

\begin{minipage}[t]{0.48\textwidth}
\centering
\textbf{(a) Prefix policy controls}
\vspace{2pt}

\begin{tabular}{llr}
\toprule
Method & K high-precision rule & MBPP \\
\midrule
KIVI K/V & None; all low-bit & 26.60 \\
Fixed K rule & Attention, $p=4$ & 25.60 \\
Fixed K rule & Attention, $p=2$ & 25.20 \\
\method{} & Selected per layer & \textbf{26.60} \\
\bottomrule
\end{tabular}
\end{minipage}
\hfill
\begin{minipage}[t]{0.48\textwidth}
\centering
\textbf{(b) Dual policy controls}
\vspace{2pt}

\begin{tabular}{llr}
\toprule
Method & K high-precision rule & MBPP \\
\midrule
Cache only & -- & 25.80 \\
KIVI K/V & None; all low-bit & 25.60 \\
Fixed K rule & Attention, $p=2$ & 25.20 \\
Fixed K rule & Attn.$\times$Q-error, $p=2$ & 26.00 \\
\method{} & Selected per layer & \textbf{26.80} \\
\bottomrule
\end{tabular}
\end{minipage}

\caption{Task-level LLaDA MBPP policy controls under 4-bit K/V cache quantization. Scores are pass@1. The $p$ value
denotes how many K tokens per token group are kept in high precision. Fixed K-rule rows use the same rule in every
layer, while \method{} selects this rule separately for each layer using trajectory calibration. Dual rows use the same generation
length, block length, and cache mode as Table~\ref{tab:llada-gen256-quant}.}
\label{tab:llada-mbpp-ablation}
\end{table*}

\paragraph{Task-level K-policy controls.}
Table~\ref{tab:llada-mbpp-ablation} compares inference-time K policies while keeping K/V quantization active. Fixed
attention-based rules remain below \method{} in both cache modes, even when they spend more K precision. In prefix
cache, \method{} matches the KIVI K/V and cache-only MBPP score, whereas the fixed attention rules reduce the score.
In dual cache, the selected policy improves over both KIVI K/V and the best fixed rule in the table. The result does
not claim that the K policy alone beats every KIVI K/V baseline; it shows that a single layer-independent
high-precision rule is insufficient, and that policy selection must depend on the layer and cache mode.

\begin{table}[t]
\centering
\small
\setlength{\tabcolsep}{5pt}
\begin{tabular}{lrrrrr}
\toprule
& \multicolumn{4}{c}{Win rate (\%)} & \\
\cmidrule(lr){2-5}
Mode & L2 & KL & Top-1 & Top-$K$ & Eff. bits \\
\midrule
Prefix & 96.4 & 88.9 & 72.9 & 85.9 & 4.77 \\
Dual & 94.9 & 90.0 & 71.7 & 82.5 & 4.60 \\
\bottomrule
\end{tabular}
\caption{Trajectory-level win rates for the compiled K-side policy on LLaDA 4-bit calibration traces. Each win rate is
the percentage of paired records improved over unprotected K quantization; L2 and KL are converted to improvements by
lower distance/divergence, while Top-1 and Top-$K$ use higher agreement/recall.}
\label{tab:policy-diagnostic}
\end{table}

\paragraph{Trajectory-level diagnostics.}
Table~\ref{tab:policy-diagnostic} evaluates the same policy on reconstructed denoising states not used for task
scoring. Each paired record compares the selected policy with the baseline that quantizes all K tokens to low bit-width.
The policy improves centered-score distance on 96.4\%/94.9\% of prefix/dual records and attention KL on
88.9\%/90.0\%, while also improving top-1 key agreement and top-$K$ key recall on most records. These diagnostics are
not downstream task scores, but they check the mechanism that motivates the method: the selected high-precision K
entries should preserve both output scores and the attention choices induced by the reused cache. The effective K
bit-widths are 4.77 in prefix cache and 4.60 in dual cache, below the 4.9-bit policy cap, so the diagnostic gain is not
explained only by a large precision increase. Together with the task-level controls, this indicates that the benefit
comes from choosing which K entries to protect, not merely from adding a uniform high-precision reserve.
\section{Related Works}
\label{sec:related-work}

\paragraph{Diffusion language models and fast inference.}
Diffusion and masked language models generate text through iterative denoising rather than causal next-token
prediction~\citep{lou2023discrete,sahoo2024simple,shi2024simplified}. Recent systems such as LLaDA and Dream show that
bidirectional masked denoising can support general language, math, and code tasks at LLM scale
~\citep{nie2026large,ye2025dream}. Fast-dLLM, dLLM-Cache, and Block Diffusion reduce the repeated full-sequence
computation required by iterative denoising through block-wise generation, cache reuse, adaptive updates, or parallel
token sampling~\citep{wu2025fast,liu2025dllm,arriola2025block}. These acceleration methods differ in schedule, but
they share the issue studied in this paper: reused KV states are approximate because bidirectional token
representations can change after later tokens are revealed. \method{} is orthogonal to these systems and quantizes the
reused cache states they produce.

\paragraph{KV cache quantization.}
LLM compression has been extensively studied through post-training weight and activation quantization, including
methods such as SmoothQuant, GPTQ, and AWQ~\citep{xiao2023smoothquant,frantar2022gptq,lin2024awq}. KV cache
quantization targets a different bottleneck: the activation memory that grows with context length, batch size, and
generation length during inference. KIVI and KVQuant are representative AR KV-cache quantizers, using asymmetric K/V
granularity, pre-RoPE key quantization, non-uniform datatypes, and outlier handling
~\citep{liu2024kivi,hooper2024kvquant}. These methods assume cached prefix states are exact reference states under
causal attention. In dLLMs, the reused cache is itself an approximation to full recomputation. We therefore adapt KIVI
and KVQuant as baselines, but evaluate quantization with behavior-level metrics and introduce trajectory-calibrated
K-side quantization under stale-cache mismatch.

\section{Conclusion}
\label{sec:conclusion}

We studied KV cache quantization for diffusion language models and show that cache reuse shifts the objective from cache reconstruction to behavior preservation under reuse-induced error. \trace{} calibrates denoising trajectories, and \method{} uses these states to select K-side policies with conservative V-side quantize-dequantize. Results on LLaDA and Dream, together with LLaDA trajectory diagnostics, support \method{} as a reproducible dLLM cache-quantization algorithm.
\section{Limitations}
\label{sec:limitations}

The evaluation covers two representative diffusion language models, two Fast-dLLM cache modes, four tasks, and both
4-bit and 2-bit KV-cache settings. It does not exhaust all decoding schedules, sequence lengths, batch sizes, or future
dLLM architectures. These extensions are natural follow-up settings for calibration and policy selection rather than
changes to the core cache-reuse error studied here.

\method{} spends its adaptive budget on K states because the method is designed to preserve attention behavior under
cache reuse. The V cache is therefore kept on a simple low-bit quantize-dequantize path. More specialized V-side policies may provide
additional gains, especially in regimes where value reconstruction rather than attention stability becomes the dominant
factor, but they are orthogonal to the main K-side behavior-preservation result. Finally, this paper reports task
quality and effective-bit accounting. End-to-end serving speed depends on fused low-bit attention kernels and runtime
integration, as in prior LLM serving and quantized-serving systems~\citep{kwon2023efficient,lin2025qserve}.


\bibliography{custom}

\appendix

\clearpage
\section{Additional Method Details}
\label{sec:appendix}

\subsection{Experimental Setup}
\label{app:experimental-setup}

We follow the Fast-dLLM evaluation protocol~\citep{wu2025fast} while adding KV-cache quantization. The evaluated models are
\texttt{GSAI-ML/LLaDA-8B-Instruct}~\citep{nie2026large} and
\texttt{Dream-org/Dream-v0-Base-7B}~\citep{ye2025dream}. The benchmark suite is shared across both models:
GSM8K~\citep{cobbe2021gsm8k}, MATH~\citep{hendrycks2021measuring}, HumanEval~\citep{chen2021humaneval}, and
MBPP~\citep{austin2021mbpp}. We use the lm-eval task definitions~\citep{eval-harness} with 5-shot GSM8K, 4-shot MATH,
0-shot HumanEval, and 3-shot MBPP. Experiments are run on NVIDIA H200 140GB GPUs.

All reported task runs use generation length 256, block length 32, and confidence-threshold parallel decoding with
threshold 0.9. Prefix and dual cache modes are evaluated separately. The no-cache row is the original diffusion
inference reference; prefix-cache and dual-cache rows are the direct cache-only references for the corresponding
quantized methods. All quantized rows use both 4-bit and 2-bit KV-cache settings and quantize both K and V unless
otherwise stated.

For adapted KIVI~\citep{liu2024kivi}, adapted KVQuant~\citep{hooper2024kvquant}, and \method{}, implementation details
are provided in Appendix~\ref{app:baseline-adaptation}. For \method{}, we use group size 32, the standard contiguous
value-side QDQ path, and a fixed K-token budget policy selected from \trace{} diagnostics. Online \method{} inference uses attention scores computed during
the block-initial forward pass and the fixed policy table; it does not use full-recompute statistics at evaluation time.

The calibration corpus is a compact mixed dataset constructed from C4~\citep{raffel2020exploring},
UltraChat~\citep{ding2023enhancing}, ARC~\citep{clark2018think}, and
CodeSearchNet-Python~\citep{husain2019codesearchnet}. The
default collection script saves 1000 C4 records, 500 UltraChat records, 300 ARC records, and 200 CodeSearchNet records
with a minimum length filter. For trajectory calibration, we use the chat-style \texttt{mixed\_chat} loader: it samples
prompt-answer pairs from the instruction, QA, and code buckets with weights 0.50, 0.25, and 0.25, respectively, applies
the model chat template to prompts when available, appends EOS to answers when available, and truncates answers to a
multiple of the block length. \method{} policy fitting uses eight calibration trajectory samples per model, cache mode,
and bit width.

\subsection{Responsible Research and Reproducibility Details}
\label{app:responsible-research}

\paragraph{Artifacts and intended use.}
This paper studies inference-time KV-cache quantization. We use publicly released pretrained checkpoints,
\texttt{GSAI-ML/LLaDA-8B-Instruct} and \texttt{Dream-org/Dream-v0-Base-7B}, and do not train new base models. The model
cards list MIT for LLaDA and Apache-2.0 for Dream; the LLaDA card reports 8B parameters, and the Dream checkpoint is a
7B-class diffusion language model whose current model card reports an 8B parameter scale. We use Fast-dLLM, KIVI,
KVQuant, Hugging Face Transformers, Accelerate, and \texttt{lm-eval}; KIVI and KVQuant are modified only to run through
the same Fast-dLLM cache wrapper described in Appendix~\ref{app:baseline-adaptation}. Any released code or policy files
are intended for research evaluation of dLLM cache compression. We do not redistribute pretrained model weights,
benchmark datasets, or the public calibration corpora.

\paragraph{Data, domains, and splits.}
All evaluation benchmarks are public English-language or Python-code benchmarks used through \texttt{lm-eval}. We
evaluate on the official test/evaluation sets: GSM8K has 1,319 test examples, HumanEval has 164 programming tasks,
MATH has 5,000 test examples, and MBPP has 500 test examples in the evaluated split. Few-shot settings are reported in
Appendix~\ref{app:experimental-setup}. Calibration uses English text, instruction/QA examples, and Python code from
C4, UltraChat, ARC, and CodeSearchNet-Python. The calibration corpus is used only to construct prompt-answer pairs and
reconstructed denoising trajectories; benchmark test labels are not used for policy fitting.

\paragraph{Privacy, offensive content, and data release.}
We do not collect new human-subject data, run new annotation studies, infer protected attributes, or scrape new data
sources. We rely on the public releases and documentation of the original datasets and do not perform an additional
manual PII or offensive-content audit. Calibration artifacts may contain tokenized or decoded snippets from the source
corpora, so they should be treated as derived research artifacts subject to the original dataset terms and should not
be redistributed without checking those terms. The task benchmarks and calibration sources may contain naturally
occurring web, instruction, or code text; our method does not attempt to filter or classify such content.

\paragraph{Experimental reporting.}
The reported task results are single-run inference evaluations under fixed decoding and quantization settings, not
averages over random seeds. Table averages are arithmetic means over the four reported tasks. The main task runs use
\texttt{transformers==4.49.0}, \texttt{lm\_eval==0.4.8}, and \texttt{accelerate==0.34.2}. The main experiments and
diagnostics ran on NVIDIA H200 140GB GPUs. The study did not train or fine-tune models; computation came from
calibration, diagnostics, and benchmark inference. Exact aggregate GPU-hours across all exploratory diagnostics were
not centrally logged, so we report the available infrastructure, model sizes, benchmark sizes, and fixed
hyperparameters rather than an exact total GPU-hour budget.

\paragraph{Risks and AI assistance.}
The method may reduce the cost of serving diffusion language models, which can improve accessibility but can also lower
the cost of misuse inherited from the underlying models. The method does not provide new safety filtering, privacy
protection, or defenses against malicious generation. Code-generation evaluation can execute generated Python programs,
so it should be run only in an appropriate sandbox. AI assistants were used during paper preparation for codebase
navigation, editing support, checklist organization, and drafting candidate wording. They were not treated as authors
and did not make final research or writing decisions; all included content was reviewed by the authors.

\subsection{KIVI and KVQuant Adaptations}
\label{app:baseline-adaptation}

We port KIVI and KVQuant into the same Fast-dLLM cache interface used by \method{}. All quantizers wrap the model with a
common cache wrapper: before a forward pass, quantized \texttt{past\_key\_values} are dequantized; after a cache-producing
forward pass, the returned K/V cache is quantized and stored. The wrapper receives the current block range and cache
mode, so the same quantizer code is used for prefix and dual cache reuse.

For KIVI, we keep the asymmetric granularity of the original method: K is quantized along the token dimension with
channel-wise scales, and V is quantized along the channel dimension with token-wise scales. The original AR decoding
implementation maintains residual high-precision tails during generation. In our Fast-dLLM adaptation, cache states are
stored at block-level reuse points, so the main experiments disable the residual tail with
\texttt{residual\_length=0} and use group size 32. This makes the KIVI baseline a direct low-bit quantization of the
reused dLLM cache rather than a separate AR streaming-cache policy.

For KVQuant, we fit a bit-width-specific calibration state from collected Fast-dLLM KV cache traces. The calibration
script stores native cache samples and, when enabled, Fisher-style activation-gradient weights. It also reshapes Dream's
3D cache layout into the common \texttt{[batch, heads, tokens, dim]} layout used by the calibration code. The deployed
quantizer loads the fitted per-layer key/value state, applies non-uniform lookup quantization, keeps sparse outlier
handling enabled with sparsity threshold 0.99, and sets \texttt{first\_few\_fp16=0}. KVQuant uses pre-RoPE cache keys
when the model exposes that cache-position mode; the other reported quantizers use the native post-RoPE cache path.

\subsection{Quantization Granularity}
\label{app:quant-granularity}

\method{} follows the K/V granularity used by representative AR KV-cache quantizers~\citep{liu2024kivi,hooper2024kvquant}.
Key activations show channel-wise structure across tokens, while value activations are better represented by token-wise
channel grouping. Accordingly, K is quantized with channel-wise scales over token groups, and V is quantized with
token-wise scales over channel groups. The dLLM-specific part of \method{} is the policy that decides which K-cache
tokens stay in high precision, not a new basic QDQ format.

\begin{figure*}[t]
  \centering
  \begin{minipage}{0.24\textwidth}
    \centering
    \includegraphics[width=\linewidth]{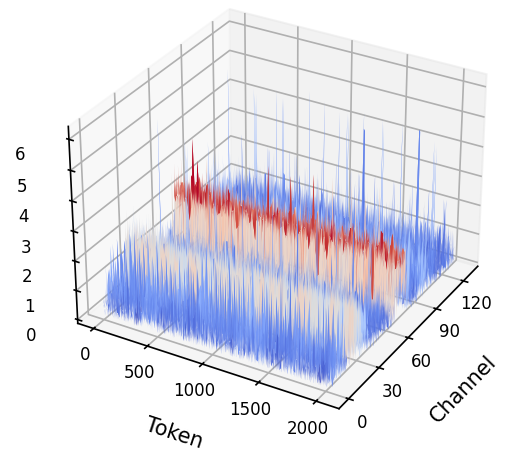}\\[-1mm]
    \footnotesize (a) Key cache, example 1
  \end{minipage}
  \hfill
  \begin{minipage}{0.24\textwidth}
    \centering
    \includegraphics[width=\linewidth]{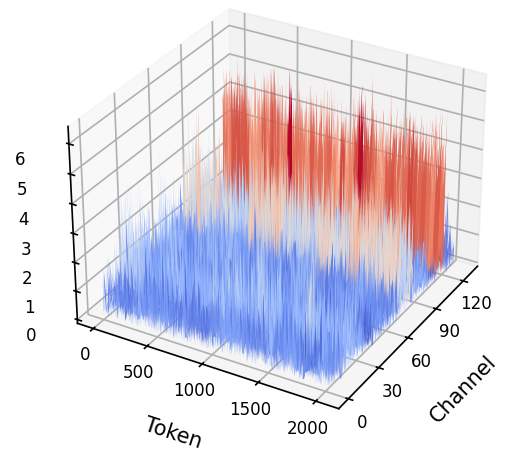}\\[-1mm]
    \footnotesize (b) Key cache, example 2
  \end{minipage}
  \hfill
  \begin{minipage}{0.24\textwidth}
    \centering
    \includegraphics[width=\linewidth]{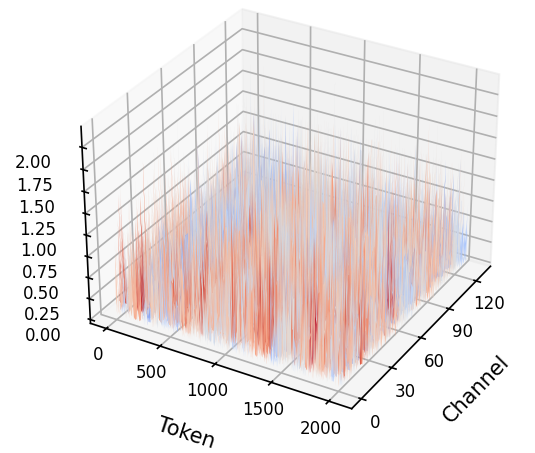}\\[-1mm]
    \footnotesize (c) Value cache, example 1
  \end{minipage}
  \hfill
  \begin{minipage}{0.24\textwidth}
    \centering
    \includegraphics[width=\linewidth]{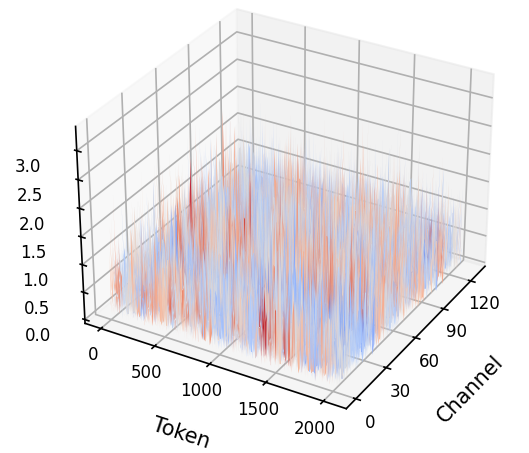}\\[-1mm]
    \footnotesize (d) Value cache, example 2
  \end{minipage}
  \caption{Representative K/V cache distributions used to choose quantization granularity.}
  \label{fig:kv-granularity}
\end{figure*}

\subsection{\trace{} Flow Construction}
\label{app:trace-flow}

\trace{} assigns answer tokens only. Prompt tokens are fixed context and are never flow items. For an answer token
$y_i$, let $\mathcal{W}_i$ be the legal reveal window induced by its block. Given the score matrix $S=[s_{i,t}]$, the
assignment solved in each refinement round is
\[
  \begin{aligned}
  \max_A\quad
  &\sum_i \sum_{t\in\mathcal{W}_i} A_{i,t}s_{i,t}\\
  \text{s.t.}\quad
  &\sum_{t\in\mathcal{W}_i} A_{i,t}=1 \quad \forall i,\\
  &\sum_i A_{i,t}=k_t \quad \forall t,\\
  &A_{i,t}\in\{0,1\}.
  \end{aligned}
\]
The corresponding min-cost flow graph has a source node, one node for each answer token, one node for each reveal
step, and a sink node. Source-to-token edges have capacity one and zero cost. Token-to-step edges exist only when
$t\in\mathcal{W}_i$, have capacity one, and have cost $-s_{i,t}$. Step-to-sink edges have capacity $k_t$ and zero
cost. The total flow equals the number of answer tokens. This construction enforces that every answer token is revealed
once and that every reveal step receives its prescribed budget.

The initializer only needs to satisfy the same feasibility constraints. The current implementation uses a deterministic
schedule-consistent initializer so that calibration artifacts are reproducible. We run a small fixed number of
rollout-refinement rounds and stop early when the assignment no longer changes. Additional sensitivity results for
initialization and refinement depth will be reported with the final calibration ablations.

\subsection{Policy Diagnostics and Selection}
\label{app:policy-selection}

Diagnostics compare each candidate against a baseline that quantizes all K tokens in the reused cache to low bit-width.
Each paired record uses the same cache mode, sample index, block, step, and layer. Lower-is-better metrics include
centered score distance and attention KL. Higher-is-better metrics include top-1 key agreement, top-$K$ key recall, and
score cosine. Each metric is converted to a signed improvement over this baseline before policy selection.

The default candidate set uses three token scores available at inference time: attention from the block-initial forward pass, K
quantization error, and the normalized product of the two. The high-precision K-token budget per group is chosen from
$\{0,1,2,4\}$. The current selection
weights are $1.0$ for centered score distance, $0.001$ for attention KL, $0.10$ for top-1 agreement, and $0.05$ for
top-$K$ recall. A candidate is accepted only when it improves the weighted score, has sufficient paired records, passes
a centered-score win-rate gate, and stays within the effective K bit-width cap. The main experiments use a win-rate
gate of $0.5$ and an effective K bit-width cap of $4.9$.

The full paper will include the selected per-layer policies for each reported model, cache mode, and bit-width. This
table is omitted from the main text because it is long, but it is part of the fixed experimental configuration rather
than a task-specific tuning step.

\subsection{Online Quantization Details}
\label{app:online-q}

K-cache quantization uses contiguous groups along the token dimension. For each group, the selected policy ranks tokens
by the chosen score and keeps the top-$p$ tokens in high precision. The remaining tokens are quantized with low-bit
min-max QDQ. The high-precision tokens are excluded from min and max estimation so that outliers do not enlarge the
low-bit quantization interval for the remaining tokens. If $p=0$, the operator reduces to ordinary KIVI-style K
quantization.

V-cache quantization uses the standard value-side low-bit QDQ path with token-wise scales over channel groups. The
submitted \method{} does not use residual tails, clipping, or learned value-side policies. This is an intentional
scope choice for the submitted method. Value-side policy variants showed weaker transfer in our diagnostics, so the
method focuses on deciding which K tokens remain in high precision.

\subsection{Additional Results to Report}
\label{app:additional-results}

The camera-ready appendix should include the following supporting material:

\begin{itemize}
  \item Full per-layer policy tables for LLaDA and Dream.
  \item Additional K/V cache distribution plots across layers and heads.
  \item TRACE initialization and refinement-depth sensitivity.
  \item Effective-bit-width and high-precision K-token budget accounting. Runtime, throughput, and peak-memory measurements
  are outside the current experimental claims because they require the optimized-kernel implementation.
  \item Calibration-sample sensitivity and per-layer policy visualizations.
  \item Appendix-only V-side diagnostics, including fixed channel grouping, range-scale, and clipping controls.
  \item Complete 4-bit and 2-bit tables for prefix and dual cache modes.
\end{itemize}

\subsection{Additional Cache Visualizations}
\label{app:block-timestep-visualization}

We additionally visualize K/V cache activations at several denoising steps within the same active block. These plots
are qualitative checks rather than policy-selection metrics. Each surface shows the absolute value of one cache head
over token and channel axes; the shaded red planes mark the active block. The examples use layer 5 and head 0 because
they are readable for both LLaDA and Dream and avoid the very large outlier scales that appear in a few deeper Dream
layers.

\begin{figure*}[p]
  \centering
  \includegraphics[width=\textwidth]{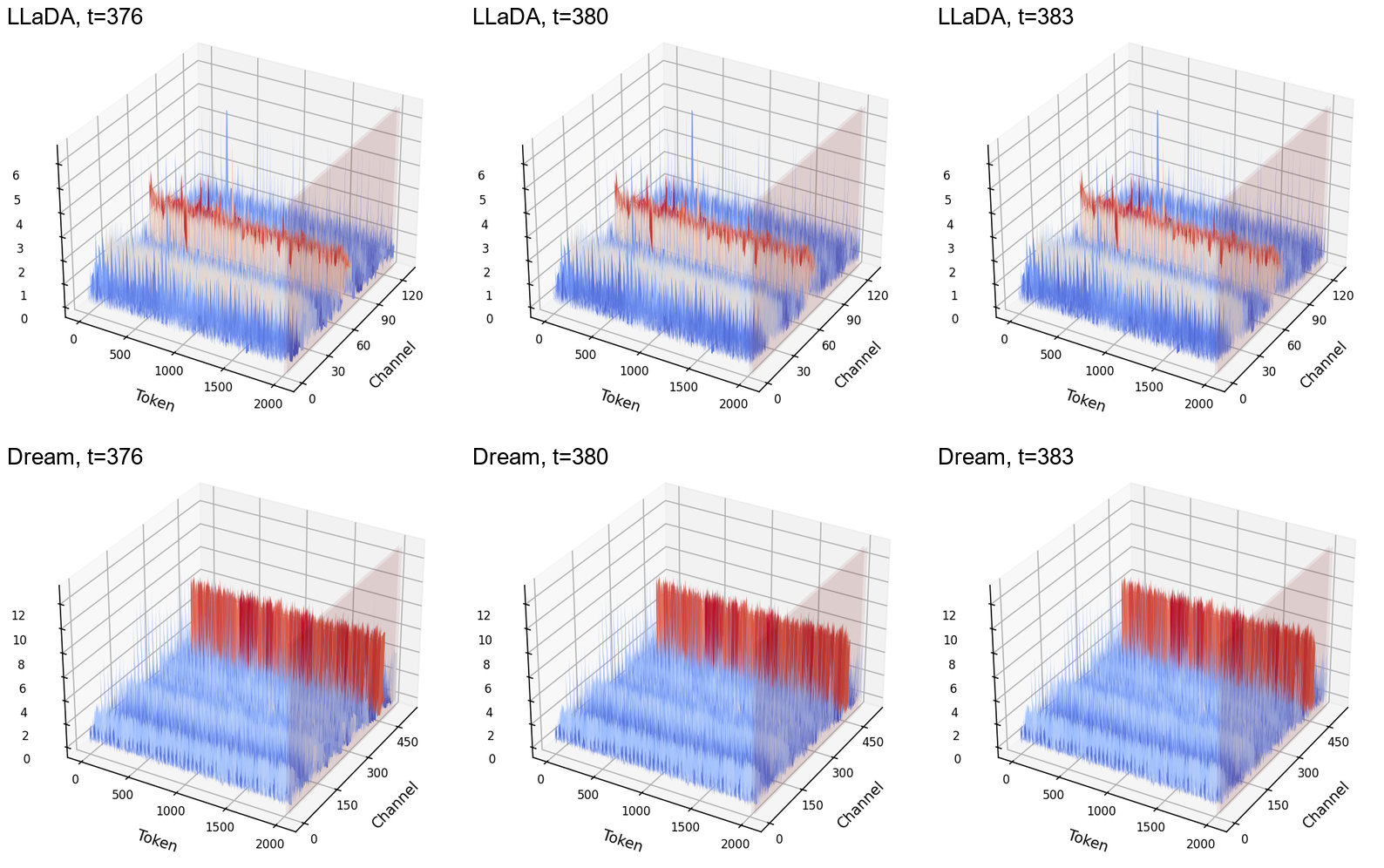}
  \caption{K-cache absolute-value surfaces across denoising steps in the same active block. The top row shows LLaDA and
  the bottom row shows Dream; columns show early, middle, and late denoising steps for block $[2016,2048)$.}
  \label{fig:block-timestep-k}
\end{figure*}

\begin{figure*}[p]
  \centering
  \includegraphics[width=\textwidth]{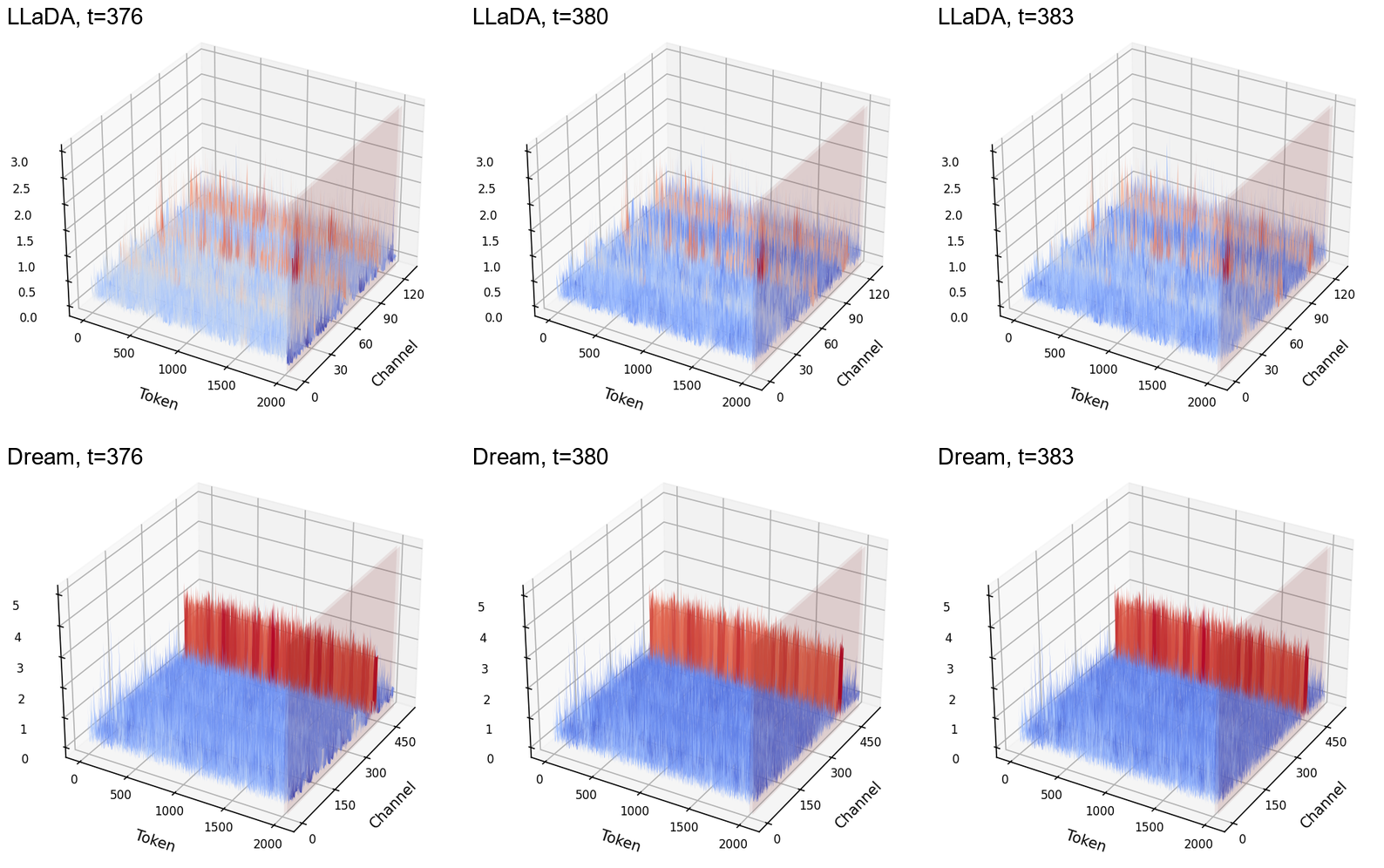}
  \caption{V-cache absolute-value surfaces for the same layer, head, block, and denoising steps as
  Figure~\ref{fig:block-timestep-k}.}
  \label{fig:block-timestep-v}
\end{figure*}

Figure~\ref{fig:block-timestep-k} shows K-cache surfaces. For both models, the K cache contains clear channel-aligned
high-magnitude regions inside and around the active block. Figure~\ref{fig:block-timestep-v} shows the corresponding
V-cache surfaces, which are less sharply concentrated for LLaDA but still structured for Dream. The visual difference
between K and V is consistent with using different K/V quantization granularity, while the model-specific scale and
pattern differences support selecting policies separately for each model and cache mode.

\end{document}